\documentclass[sigconf,natbib=true,anonymous=false]{acmart}

\AtBeginDocument{%
  \providecommand\BibTeX{{%
    \normalfont B\kern-0.5em{\scshape i\kern-0.25em b}\kern-0.8em\TeX}}}

\copyrightyear{2024}
\acmYear{2024}
\setcopyright{acmlicensed}\acmConference[SIGIR '24]{Proceedings of the 47th International ACM SIGIR Conference on Research and Development in Information Retrieval}{July 14--18, 2024}{Washington, DC, USA}
\acmBooktitle{Proceedings of the 47th International ACM SIGIR Conference on Research and Development in Information Retrieval (SIGIR '24), July 14--18, 2024, Washington, DC, USA}
\acmDOI{10.1145/3626772.3657816}
\acmISBN{979-8-4007-0431-4/24/07}

\makeatletter
\gdef\@copyrightpermission{
  \begin{minipage}{0.3\columnwidth}
   \href{https://creativecommons.org/licenses/by/4.0/}{\includegraphics[width=0.90\textwidth]{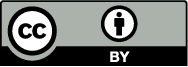}}
  \end{minipage}\hfill
  \begin{minipage}{0.7\columnwidth}
   \href{https://creativecommons.org/licenses/by/4.0/}{This work is licensed under a Creative Commons Attribution International 4.0 License.}
  \end{minipage}
  \vspace{5pt}
}
\makeatother

\acmSubmissionID{fp8362}

\usepackage[utf8]{inputenc} 
\usepackage[T1]{fontenc}    
\usepackage{hyperref}       
\usepackage{url}            
\usepackage{booktabs}       
\usepackage{amsfonts}       
\usepackage{nicefrac}       
\usepackage{microtype}      
\usepackage{xcolor}         

\usepackage{graphicx}
\usepackage{caption}
\usepackage{subcaption}

\usepackage{ulem}
\usepackage{amsmath}
\usepackage{multirow}

\usepackage[utf8]{inputenc}
\usepackage{xspace}
\newcommand*{\eg}{e.g.\@\xspace}
\newcommand*{\ie}{i.e.\@\xspace}

\newcommand*{\wrt}{w.r.t\@\xspace}

\usepackage{xcolor}
\usepackage{soul}

\usepackage{amsmath}

\usepackage{siunitx}

\newcommand{\OURS}{SRT}

\begin{document}

\title{Scaling Sequential Recommendation Models with Transformers}

\author{Pablo Zivic}
\orcid{0000-0002-9872-6007}
\affiliation{%
 \institution{Mercado Libre Inc.}
 \streetaddress{Posta 4789}
 \city{Buenos Aires}
 \country{Argentina}
 \postcode{C1430EKH}
}
\email{pablo.rzivic@mercadolibre.com}

\author{Hernan Vazquez}
\orcid{0009-0002-4363-3193}
\affiliation{%
 \institution{Mercado Libre Inc.}
 \streetaddress{Posta 4789}
 \city{Buenos Aires}
 \country{Argentina}
 \postcode{C1430EKH}
}
\email{hernan.vazquez@mercadolibre.com}

\author{Jorge S\'anchez}
\orcid{0000-0002-0150-7263}
\affiliation{%
 \institution{Mercado Libre Inc.}
 \streetaddress{Humberto 1o 630}
 \city{C\'ordoba}
 \country{Argentina}
 \postcode{X5000}
}
\email{jorge.sanchez@mercadolibre.com}

\renewcommand{\shortauthors}{Pablo Zivic, Hernan Vazquez, \& Jorge Sánchez}
\begin{abstract}
Modeling user preferences has been mainly addressed by looking at users' interaction history with the different elements available in the system. Tailoring content to individual preferences based on historical data is the main goal of sequential recommendation.
The nature of the problem, as well as the good performance observed across various domains, has motivated the use of the transformer architecture, which has proven effective in leveraging increasingly larger amounts of training data when accompanied by an increase in the number of model parameters. This scaling behavior has brought a great deal of attention, as it provides valuable guidance in the design and training of even larger models.
Taking inspiration from the scaling laws observed in training large language models, we explore similar principles for sequential recommendation. 
Addressing scalability in this context requires special considerations as some particularities of the problem depart from the language modeling case. These particularities originate in the nature of the content catalogs, which are significantly larger than the vocabularies used for language and might change over time. In our case, we start from a well-known transformer-based model from the literature and make two crucial modifications. 
First, we pivot from the traditional representation of catalog items as trainable embeddings to representations computed with a trainable feature extractor, making the parameter count independent of the number of items in the catalog. 
Second, we propose a contrastive learning formulation that provides us with a better representation of the catalog diversity. We demonstrate that, under this setting, we can train our models effectively on increasingly larger datasets under a common experimental setup.
We use the full Amazon Product Data dataset, which has only been partially explored in other studies, and reveal scaling behaviors similar to those found in language models. Compute-optimal training is possible but requires a careful analysis of the compute-performance trade-offs specific to the application. 
We also show that performance scaling translates to downstream tasks by fine-tuning larger pre-trained models on smaller task-specific domains. Our approach and findings provide a strategic roadmap for model training and deployment in real high-dimensional preference spaces, facilitating better training and inference efficiency. 
We hope this paper bridges the gap between the potential of transformers and the intrinsic complexities of high-dimensional sequential recommendation in real-world recommender systems.
Code and models can be found at {\small\url{https://github.com/mercadolibre/srt}}.

\end{abstract}

\ccsdesc[500]{Information systems}
\ccsdesc[500]{Information systems~Personalization}
\ccsdesc[500]{Information systems~Recommender systems}
\ccsdesc[500]{Information systems~Personalization}

\keywords{Sequential Recommendation, Scaling Laws, Transformers, Transfer Learning}


\maketitle

\section{Introduction}

A recommendation system aims to provide users with content that fits their preferences and interests. Classical methods have explored building static models based on past user interactions to predict future ones. Pattern mining and factorization-based methods are two classical methodologies that stand as the most popular in the literature \cite{su2009survey,fang2020deep}. These models seek to capture static preferences in the interaction of users with items in a catalog. While the formulation largely simplifies the modeling of otherwise complex interaction patterns observed in the real world, the main drawback of these models relies on the assumption of static behavior patterns. In reality, user preferences are subject to a series of short- and long-term factors that are very hard to disentangle \cite{campos2014time,rafailidis2015modeling}. From a modeling perspective, user preference dynamics can be seen as latent factors that govern the observed users' behavior as they interact with the system. These interactions are diverse and depend on the nature of the actual system. For instance, the types of events that can be registered in a music streaming platform differ from those observed in an e-commerce website. Despite the complexity of the task, the driving hypothesis of modern recommendation systems is that such behavioral patterns can be captured by models that can predict future interactions from historical sequential records. 
The nature and complexity of such models have been influenced to a great extent by the success of different machine-learning models in different fields, especially those from the natural language literature. We can find solutions based on simple Recurrent Neural Networks \cite{hidasi2015session}, convolutional architectures \cite{tang2018personalized,tuan20173d}, based on Attention mechanisms \cite{li2017neural,zhang2019next} and, more recently, the Transformer \cite{vaswani2017attention,sun2019bert4rec, kang2018self}. Among them, transformer-based solutions are the most promising. This is not only due to the success of this architecture in fields beyond language modeling, such as computer vision \cite{khan2022transformers}, speech \cite{latif2023transformers}, and time-series forecasting \cite{lim2021temporal}, but also to their flexibility and good scaling behavior. Another important factor of the transformer architecture that led to its adoption as the model of choice in many applications is the availability of a pre-fitted version that can be adapted easily to more specific tasks using a fraction of the data required to train a similar model from scratch. This pre-training and fine-tuning strategy has not yet been widely adopted in the sequential recommendation literature. We believe that this obeys two main reasons: first, the data used to train recommendation models is specific to each application domain, \ie the nature of the catalog and type of events are problem-specific, making it challenging to leverage prefitted models on domains that might be closely related but not the same as those on which the model has been trained (\eg a model pre-trained on Amazon data being fine-tuned to a different catalog or e-commerce domain); second, there are particularities in the sequential recommendation problem that constraint the design of solutions that scale. 

While predicting the next token in a sentence and the next item in an interaction sequence share structural similarities, the sequential recommendation problem introduces some particularities that need special attention. For instance, in language modeling, it is common to cast the prediction task as a classification problem over a large set of tokens (subdivisions of words into finer sub-word units). Although large, the size of this vocabulary remains constrained to a manageable number (around 30K in most practical applications) that does not change over time. On the contrary, most real recommendation applications involve item sets (space of possible user preferences) that expand to massive scales, often reaching into the millions or even billions of different items \cite{covington2016deep}. Moreover, such collections may change over time as items are constantly added and removed from the catalog. These characteristics impose design constraints that must be satisfied if we are willing to take advantage of the flexibility and ease of adaptation observed by these architectures in other domains.

Perhaps the most intriguing characteristic of these models relies on their ability to leverage increasingly large amounts of data by simply growing the number of parameters accordingly. From a system design perspective, this poses new challenges around how to scale the amount of data and compute required by these models to leverage their full potential. From a practical perspective, this choice is constrained not only by the desire to get the best possible performance but also to achieve such performance within the limits of a given computational budget. In light of recent discoveries regarding scaling laws in the language \cite{kaplan2020scaling, alabdulmohsin2022revisiting} and other domains \cite{rosenfeld2019constructive, zhai2022scaling, henighan2020scaling}, recent research has provided new insights into how model performance scales with the number of parameters, the size of the datasets used for training, and the required computational budget \cite{hoffmann2022training, alabdulmohsin2023getting}. 

In this work, we explore the hypothesis that transformer-based sequential recommendation models exhibit scaling behaviors similar to those observed in other domains. Under such a hypothesis, we investigate how, within a given computational budget, optimizing the balance between model size and data size can yield improved results. To do so, we propose a generic yet scalable model that takes inspiration from other transformer-based models from the literature but lets us experiment with problems and models of different complexity. We run experiments on the full version of the widely used Amazon Product Data (APD) dataset \cite{ni2019justifying}. Our findings confirm our hypothesis, and we show how such scaling behavior can be used in practice by training larger models that, when fine-tuned, achieve a performance that surpasses more complex approaches from the literature. Our main contributions are the following:
\begin{itemize}
    \item We propose a generic transformer-based architecture that is both flexible and scalable.
    \item We show scaling laws similar to those observed in language modeling tasks. 
    \item We show that it is possible to pre-train recommendation models at scale and fine-tune them to particular downstream tasks, improving performance \wrt to similar models trained from scratch.
\end{itemize}

The paper is organized as follows: Sec.~\ref{sec:preliminaries} discusses some key aspects of sequential recommendation models in the context of scalability, Sec.~\ref{sec:framework} proposes a formulation that makes the model independent of the size of the catalog, Sec.~\ref{sec:experiments} shows experimental results. In Sec.~\ref{sec:scaling_laws} we derive analytical laws that relate the target metric with the most relevant quantities of interest from a scaling perspective. In Sec.~\ref{sec:transfer_learning} we show we can use the pre-training and fine-tuning strategy for improving recommendations. Sec.~\ref{sec:related_work} discusses related work. Finally, in Sec.~\ref{sec:conclusions} we draw some conclusions.

\section{Scalability of Sequential Recommendation Models}
\label{sec:preliminaries}

Sequential recommendation models seek to capture user interaction patterns and a possibly large collection of available \textit{items} in a catalog. Users may perform many interaction types depending on the nature of such elements: an e-commerce site, a music streaming service, a social network, and others. For instance, users might play a song, skip it, or add it to a playlist in a music recommendation context, while they can add an article to a shopping cart, buy it, add it to a wish list, etc. Although such heterogeneity in the type of interactions can be handled accordingly \cite{peng2021lime, jin2020multi, wu2022multi}, for the purpose of this study, we subsume all domain-specific cases into a more generic "user-item" interaction (\ie a user interacted with an item in some way). 
With this in mind, let $U$ denote the user base of a given platform and $I$ the collection of items they can interact with. The behavior and preference dynamics of a user $u\in U$ can be considered as embedded into the sequence of items they interacted with for a given period. Let $S_u=\{i^u_1, i^u_2, \dots i^u_n\}$ be such a sequence, where $i^u_k$ denotes the $k$-th item user $u$ interacted with. In this context, a recommendation model can be thought of as a function $f_\theta$, parameterized by $\theta$, that takes as input the interaction history encoded by $S_u$ and seeks to predict the item or items that user $u$ will interact with in the future. 

\begin{figure}[h]
    \centering
    \includegraphics[width=\linewidth]{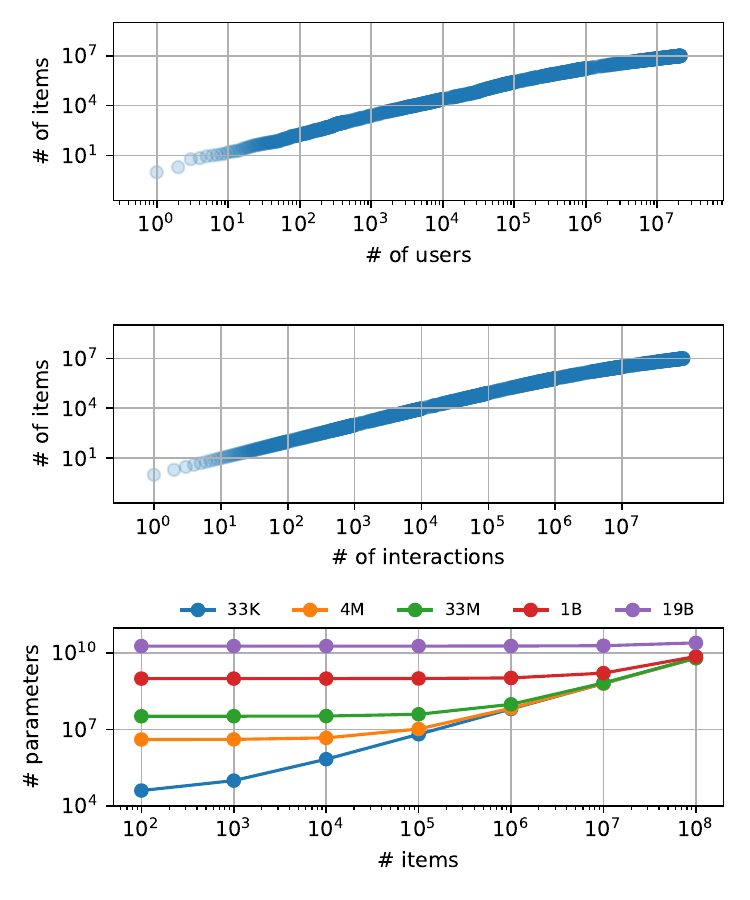}
    \caption{Increase in the catalog size induced by an increase in the number of active users (top) and interactions (middle). Increasing the number of items increases the total parameter count (bottom) in models that use trainable item embeddings for 64-dimensional embeddings. For smaller models (33K parameters), the catalog size dominates the total parameter count, while for larger models (19B parameters), the total number of trainable parameters remains stable across a wide spectrum of catalog sizes.}
    \label{fig:param-count}
\end{figure}

Given the sequential nature of the problem, we consider models based on the transformer architecture \cite{latifi2022sequential}. These models, initially proposed in the context of language modeling tasks, have proven effective in various domains. From a scaling perspective, and similarly to what happens in the natural language case, we have two clear dimensions that affect their scaling behavior: the number of model parameters, $N$, and the number of user-item interactions seen during training. 
However, as we will see next, analyzing scalability in most recommender systems proposed in the literature would also require considering the number of available items in the catalog, $|I|$. This dependency originates in the way most transformer-based approaches cast the sequential prediction task. 
In a direct translation of the next-word prediction used to train language models, recommendation transformers treat the elements of the input sequence (items from the catalog) as "tokens", \ie atomic elements whose co-occurrence patterns we try to learn from data. As in the language modeling case, the model is asked to learn the interaction patterns between sequences of such atoms and also a representation (embedding) that encodes some intrinsic aspect of each such element. This implies that, besides the actual number of parameters in the model devoted to sequential prediction, we also need to store a number of vector embeddings equal to the catalog size. 
While in the case of language the vocabulary size is relatively small, ranging between the tens to a few hundred thousand elements, the size of the catalog for a real-world recommendation model can grow dramatically, potentially reaching into the billions \cite{covington2016deep}. In such cases, the number of parameters associated with the matrix of trainable item embeddings quickly dominates the total parameter count. Such a dependency between model complexity and the size of the catalog is depicted in Fig.~\ref{fig:param-count}. 
In the bottom panel of the figure, we show an estimate of the total number of parameters for models of different sizes (expressed in terms of the total number of parameters, $N$, as a function of the size of the catalog, $|I|$. We observe that for models with a small complexity (33K and 4M parameters), the total parameter count grows linearly with the size of the catalog after a relatively short nearly flat initial regime. For larger models ($>1B$), on the other hand, the total parameter count remains stable across a wide range of $|I|$, growing only for extremely large values of $|I|$.

This phenomenon makes the analysis difficult since, by increasing the number of training sequences, we would implicitly increase the number of items interacted with by users, which would, in turn, increase the number of parameters of the model. This growth is not deterministic and depends on the diversity of items observed in the pool of sequences used for training. This behavior is illustrated in the first two panels of Fig.~\ref{fig:param-count}, where we show the growth in the number of visited catalog items induced by an increase in the number of active users (top) and number of navigation data (middle). Moreover, building a solution based on learning item embeddings worsens the cold-start problem observed in real systems \cite{lika2014facing,gope2017survey}.

In the next section, we reformulate the standard transformer-based approach to break this dependency. In this way, the complexity of the model becomes independent of $|I|$ and we can analyze the scaling behavior \wrt the variables of interest (parameter count and training set size) concisely.

\section{A Scalable Recommendation Framework}
\label{sec:framework}

We take SASRec \cite{kang2018self} as our reference model. SASRec is a transformer-based architecture that has shown competitive performance on several sequential recommendation tasks \cite{li2021lightweight} and is regarded as a strong baseline in more recent evaluations \cite{petrov2022systematic}. In SASRec, the model takes as input a sequence of user-item interactions of length $n$ and seeks to predict the item the user will interact with next. The sequence is fed into a transformer model of $L$ layers to produce an output embedding that matches a representation of the following item in the sequence. The output might also include a classification layer over the items in the catalog that induces additional complexity \cite{sun2019bert4rec}. Each item in the catalog is encoded as a trainable vector representation of size $D$, resulting in a total of $|I| \times D$ trainable parameters. As mentioned above, this dependency between the number of trainable parameters and the catalog size makes the analysis of scaling behaviors difficult due to the interplay between $N$ and $|I|$.

\begin{figure}[h]
    \centering
    \begin{subfigure}[c]{0.22\textwidth}
        \centering
        \includegraphics[width=\textwidth]{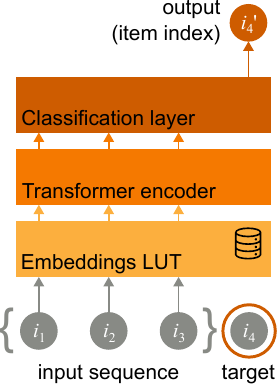}
        \label{fig:archs_a}
    \end{subfigure}
    \hspace{10pt}
    \begin{subfigure}[c]{0.22\textwidth}
        \centering
        \includegraphics[width=\textwidth]{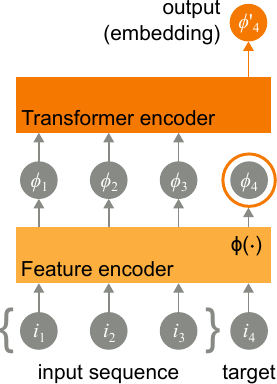}
        \label{fig:archs_b}
    \end{subfigure}
    \caption{Traditional transformer-based recommendation models (left) learn vector embeddings for all items in the catalog and access them via table lookups (LUT). The system might also include a classification layer aimed at predicting the index of the target item. On the contrary, we propose to use a fixed (and task-agnostic) feature extractor to encode the items in the catalog (right) and to predict item-to-item similarities using the output embeddings of the transformer model. We use the last element of the sequence as target and try to match its id (classification) or input embedding (regression).}
    \label{fig:archs}
\end{figure}

Figure~\ref{fig:archs} (left) illustrates this scenario, where both the input and output sequences correspond (to indices) to items in the catalog. For large catalogs, the parameter count is dominated by the matrix of input embeddings, the output prediction layer(s), or both. Given these observations, instead of considering the learning problem as a classification one, we propose reformulating it as an embedding regression task \cite{tan2016improved}, as follows. 
Given a user navigation sequence $S_u=\{i^u_1, i^u_2, \dots i^u_n\}$, we assume we can compute a $D$-dimensional vector representation for each item in the sequence. To compute such representations, we rely on a parametric mapping $\phi:I\rightarrow\mathbb{R}^D$. We denote as $\phi^u_k\equiv\phi(i^u_k)$ the representation of $k$-th item user $u$ interacted with. These embeddings are user-independent in that the representation for a given item is the same irrespective of how the user might interact with it. We compute these representations on the fly and train the full model (including the feature extraction model $\phi$) to pick among (a subset of) them the one that corresponds to the target item for an input sequence $S_u$. 
This is illustrated in Figure~\ref{fig:archs} (right), where we have replaced the embedding layer with a feature extractor that computes the item embeddings ($\phi_i, i=1,..,4$) that feed the model. The output of this model ($\phi'_4$) is used for prediction (and learning) by comparing similarities with the items in the catalog (their embeddings). This feature-based approach has shown good performance in the literature \cite{zhang2019feature, qiu2021memory}. In this case, the number of parameters associated with the computation of item embeddings is given by the number of trainable parameters in the feature extraction module $\phi$, irrespective of the number of items in the catalog.

We train our model autoregressively as follows. Given a training set of user navigation sequences of length $n$, we ask the model to predict each element of any given sequence based on the (sub)sequence of previous interactions. We optimize the following loss:
\begin{equation}
    \mathcal{L}(\theta; \mathcal{S}) = \sum_{S\in\mathcal{S}} \ell(\theta; S).
\end{equation}
Here, we omitted the superscript $u$ for the sake of clarity\footnote{Also, the information we are willing to capture relates to the preferences of users and the way they interact with the system, and not on their particular identities.}. Let $\mathcal{S}$ denote the set of all subsequences with a length of at least 2 interactions. Let us denote by $\tilde{S}=\{i_1, \dots, i_{n-1}\}$ the partial sequence containing the first $n-1$ items of $S$. Our goal is to train a model that, based on $\tilde{S}$, can rank the target $i_n$ as high as possible when compared to other candidates from the catalog. We adopt sampled Softmax \cite{wu2022effectiveness} as our choice for $\ell$: 
\begin{equation}
    \mathcal{\ell}(\theta; S) = -\log\frac{\exp \left({f_\theta(\tilde{S})}^T \phi_n/\tau \right)}{\exp \left({f_\theta(\tilde{S})}^T \phi_n/\tau\right) + \sum_{\phi\in\mathcal{N}(\tilde{S})} \exp \left({f_\theta(\tilde{S})}^T \phi/\tau\right)} .
    \label{eq:loss}
\end{equation}

Here, $f_\theta$ denotes the model we are trying to fit, $\phi_n$ the representation of the target item $i_n$ computed by the feature extraction module, $\tau$ is a temperature parameter that controls the softness of the positive and negative interactions, and $\mathcal{N}(\tilde{S}) \subset I$ is a set of negatives whose cardinality is to be set. In practice, we apply the logQ correction proposed in \cite{yi2019sampling} to the logits in Eq.~\eqref{eq:loss} to correct for the bias introduced by the negative sampling distribution.
In the rest of the paper, we refer to our transformer-based model and learning formulation as \textit{Scalable Recommendation Transformer (\OURS)}.

The framework introduced above is motivated by the need to set up a competitive yet simple baseline that scales well \wrt the quantities we identified as the most relevant from a scaling perspective, namely the number of trainable parameters and the number of samples (or interactions) observed during training.

Eq.~\eqref{eq:loss} can be seen as an approximation to a cross-entropy loss over the items in the catalog, where we contrast against a subset of the possible items. This corresponds to a generalization of the loss used in SASRec or BERT4Rec, where $\mathcal{N}(\tilde{S})$ is constrained to a single sample draw at random. Note that by drawing samples at random, we take the risk of contrasting against uninformative samples that are easily distinguishable from the positive ones. On the other hand, if we choose an elaborate negative sampling methodology, we might end up adding a non-negligible computation overhead to an otherwise simple model. 
Moreover, sampling hard negatives might induce biases that correlate with the catalog size \cite{liu2021contrastive}, adding a degree of variability that is difficult to isolate. In our case, we opt to sample negatives from the item popularity distribution (\ie items with which users interacted the most are sampled more frequently). This strategy is competitive and has a small footprint on the overall computations. From now on, we denote our models as \OURS-X, where X is the number of negatives used to compute the loss in Eq.~\eqref{eq:loss}.

Besides the advantages of the proposed formulation regarding scalability, an additional advantage of our model is the ability to work with non-static catalogs. Adding and deleting items dynamically from a catalog (due to policy infringements, product stockout, new trends, outdated information, etc) is commonplace in most practical applications. Building sequential recommendation models based on fixed item sets brings many concerns regarding the usability and maintainability of the system over time. These concerns might hinder a wider adoption of these types of approaches.

Before delving deeper into scalability, which is the primary goal of our work, we first show that our formulation achieves competitive performance compared to other transformer-based formulations. Table~\ref{tab:amzn_baselines} compares the performance of our model against other popular methods from the literature on the Beauty and Sports subsets of the Amazon Review Data \cite{mcauley2015image} benchmark. Details of the dataset, metrics, and evaluation protocols are provided in Sec.~\ref{sec:experiments}. 

\begin{table}[!htp]\centering
\caption{Performance comparison between two reference models and our formulation under the NDCG@5 metric for the Amazon Beauty and Sports datasets. 
}
\begin{tabular}{lcc}\toprule
&Beauty &Sports \\\midrule
BERT4Rec \cite{sun2019bert4rec} &0.0219 &0.0143 \\
SASRec \cite{kang2018self} &0.0241 &0.0135 \\
SASRec-CE &0.0314 &0.0170 \\\midrule
SRT-10 &0.0235 &0.0138 \\
SRT-100 &0.0318 &0.0171 \\
SRT-300 &0.0340 &0.0180 \\
SRT-1K &0.0366 &0.0195 \\
\bottomrule
\end{tabular}
\label{tab:amzn_baselines}
\end{table}
We consider different versions of our model trained using 10, 100, 300, and 1K negatives, respectively, and compare them against the popular SASRec \cite{kang2018self} and BERT4Rec \cite{sun2019bert4rec} models. 
For these models, we show the metrics reported by \citet{chen2022intent} for compatibility of evaluation methodology.
We also report a variation of SASRec in which we replaced the binary cross-entropy with a full cross-entropy loss over the items in the catalog. We name this variant as SASRec-CE. There are several observations to be made. First, we see that our best model outperforms all other alternatives, even when, at its core, the underlying models are very similar. Second, the negative sampling strategy is crucial in getting good performance.
This is interesting since we can see the multiclass cross-entropy loss as a measure that contrasts each positive against all negatives (full catalog) and, in that sense, can be seen as a limiting case for our contrastive formulation. Increasing the number of samples beyond this value becomes impractical as it involves computing embeddings for additional $XB$ samples, with $B$ the size of the training batch. We believe these results validate the overall formulation and set a strong baseline model for scalability analysis.

\section{Experiments}
\label{sec:experiments}

This section discusses our experimental setup in the context of standard practices observed in the literature. We then show and discuss results on scalability and optimal compute allocation. Finally, we show fine-tuning results that compete favorably with other methods from the literature.

\subsection{Evaluation Protocol}

\begin{figure}[h]
    \centering
    \includegraphics[width=0.95\linewidth]{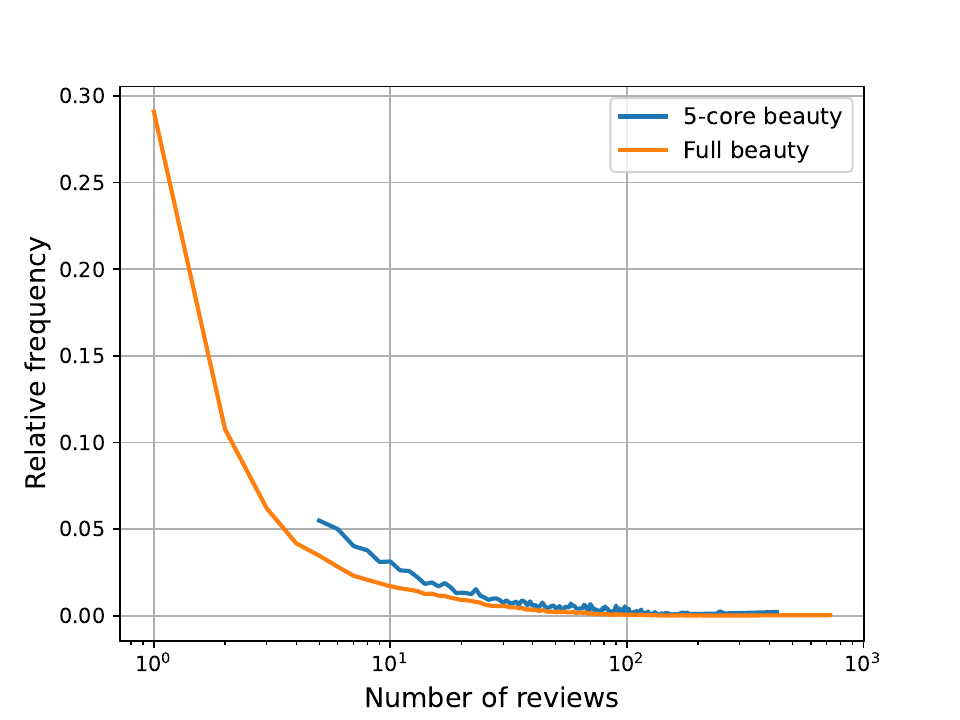}
    \caption{Distribution of the number of reviews per item in the Amazon beauty dataset for the full and 5-core versions. Similar behaviors are observed in Amazon sports.}
    \label{fig:long_tail}
\end{figure}

We ran experiments on the Amazon Product Data (APD) dataset \cite{mcauley2015image,he2016ups}, a large dataset of product reviews crawled from Amazon between 1996 and 2014. The dataset consists of $82.7$ million reviews over $9.9$ million different products written by more than $21$ million users. Reviews in this dataset correspond to a subset of all purchases made in the platform during the relevant time span. Due to its size, a common practice in the literature consists of using smaller subsets of the data. For instance, "Amazon Beauty" corresponds to the subset of samples where users bought (and reviewed) an item from the "beauty" category. To avoid issues related to cold-start \cite{lika2014facing}, it is common to filter out users and products with less than five purchases. The remaining data is called a "$5$-core" dataset. These two procedures (per-category and $5$-core filtering) distort or hide some of the intrinsic characteristics of real-world recommendation problems. For instance, if we consider the "beauty" category, only $8.7\%$ of the interactions originate from items with at least $5$ purchases/reviews. This is not only a matter of scale (with most data being discarded) but a problem of deceiving evaluation, as results reported on these datasets do not necessarily extrapolate to actual real systems. Table~\ref{tab:stats} provides dataset statistics for the entire dataset, two common subsets used in the literature, and their 5-core trimmed versions. There are an order of magnitude fewer interactions and items in the $5$-core version of the dataset compared to their full counterpart. Figure~\ref{fig:long_tail} shows the distribution of reviews per item for the beauty subset of APD for both the full and 5-core versions. As the figure shows, trimming the dataset reshapes the original problem into modeling a long-tail phenomenon, disregarding the rich and more relevant aspects of real user interactions. 
\begin{table}
    \centering    
    \caption{Statistics for the full and 5-core trimmed versions of Amazon Product Data and the Beauty and Sports categories.}
    \begin{tabular}{lccccc}\toprule
    &\multirow{2}{*}{APD} &\multicolumn{2}{c}{Beauty} &\multicolumn{2}{c}{Sports} \\\cmidrule(l{0pt}r{2pt}){3-4}\cmidrule(l{2pt}r{0pt}){5-6}
    & &raw &5-core &raw &5-core \\\midrule
    \# interactions &82.7M &2.3M &198.5K &2.5M &296.3K \\
    \# users &21.2M &22.4K &22.4K &35.6K &35.6K \\
    \# items &9.9M &937.9K &12.1K &993.6K &18.4K \\
    \# iter/item (avg) &8.4 &2.4 &16.4 &2.5 &16.1 \\
    \# iter/user (avg) &3.9 &101.9 &8.9 &69.9 & 8.32 \\
    \bottomrule
    \end{tabular}
    \label{tab:stats}
\end{table}
From the above and to analyze the scaling behavior of transformer-based recommendation models, we propose a simple strategy to generate datasets of different sizes with the same distributional information as the original (average sequence length, diversity, etc.). The strategy consists of sampling users at random and recording their interactions into a single dataset. Note that by increasing the number of active users, we account for a more extensive set of items they interact with (see Fig.~\ref{fig:param-count}). We follow standard practice and take the last item from each sequence as the target for evaluation (test), the item at position $n-1$ for validation, and leave the first $n-2$ elements for training. 

Due to the long-tail behavior of item categories, we expect larger datasets to account also for larger catalogs. This poses a challenge from an evaluation perspective since computing retrieval metrics over the whole catalog becomes infeasible. We sample a set of random negatives for each positive, as in \cite{kang2018self}. In our case, however, we sample 10K negatives for each positive instead of only 100 as in \cite{kang2018self}. By doing so, we increase the probability of sampling negatives, which are more challenging to discriminate \wrt the ground truth, while providing us with a closer approximation to the full-catalog case. In our experiments, we chose the NDCG@5 score as a performance metric as it is one of the most common metrics reported in the literature.

Following \cite{kaplan2020scaling, hoffmann2022training}, we use the number of FLOPs as a proxy for the amount of compute required to achieve a given performance for different choices of model complexity, number of training samples, and the catalog size.
    
\subsection{Model Design and Training Algorithm}

We adapt SASRec as outlined in Sec.~\ref{sec:framework}, replacing the item embedding matrix with a trainable feature encoder whose complexity is independent of the size of the catalog. Concretely, we take the title and brand of each product and tokenize them into a vocabulary of $30k$ tokens with the SentencePiece tokenizer \cite{kudo2018sentencepiece}. This way, we replace the variable-sized item embedding matrix with a fixed matrix of token embeddings. As shown in Table~\ref{tab:amzn_baselines}, these changes lead to comparable performance in standard benchmarks.
Based on this architecture, we consider different model complexities parameterized by the number of layers, $n_L$, number of attention heads per layer, $n_H$, and hidden embedding dimensionality, $d$. Table~\ref{tab:model_sizes} details the different combinations of these parameters we used in our experiments.
\begin{table}
    \centering    
    \caption{Architecture specification. Each column represents a different model parameterized by the number of layers, $n_L$, number of attention heads, $n_H$, and number of hidden embedding dimensions, $d$.}
\begin{tabular}{lcccccccc}\toprule
$n_L$ &24 &16 &8 &8 &8 &4 &2 &4 \\
$n_H$ &4 &4 &4 &4 &2 &2 &2 &2 \\
$d$ &256 &256 &256 &128 &128 &128 &128 &64 \\
\bottomrule
\end{tabular}
    \label{tab:model_sizes}
\end{table}
To train our models, we use the Adam optimizer and a one-cycle learning rate policy consisting of a linear warm-up stage and a cosine decay after one-third of the total iterations. We set the base learning rate to $1e-4$ and the number of epochs to $50$. We use gradient clipping (set to $1$) and a weight decay factor of $1e-5$. We base our implementations on the RecBole library \cite{recbole}.

\subsection{Scaling Model and Dataset Sizes for Optimal Compute}

In this section, we explore the relationship between the target metric and the compute resource requirements induced by different combinations of model sizes and number of training interactions. Our evaluation differs from similar studies \cite{clark2022unified,kaplan2020scaling,hoffmann2022training} in two main aspects: first, we focus on a task-specific metric instead of a more generic loss; second, we train our models over multiple epochs, thus revisiting the same training sequences multiple times during the training process. These differences originate from the particularities of the sequential recommendation problem. We also focus on a performance metric that is closely tied to the actual recommendation task (NDCG vs loss as in the language modeling case) and which is more informative from a practical standpoint.

\begin{figure*}[h]
    \centering
    \includegraphics[width=0.45\linewidth]{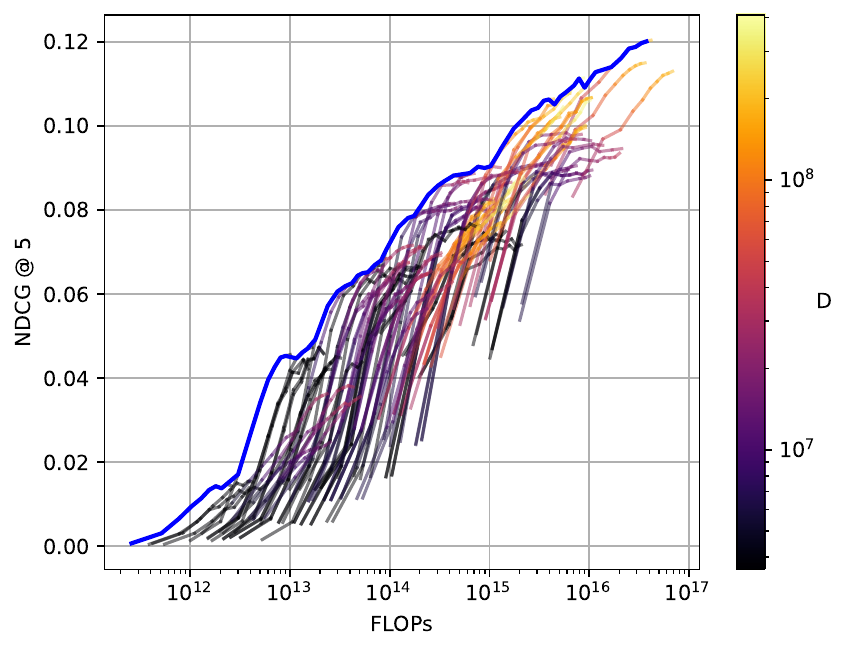}\hspace{15pt}
    \includegraphics[width=0.45\linewidth]{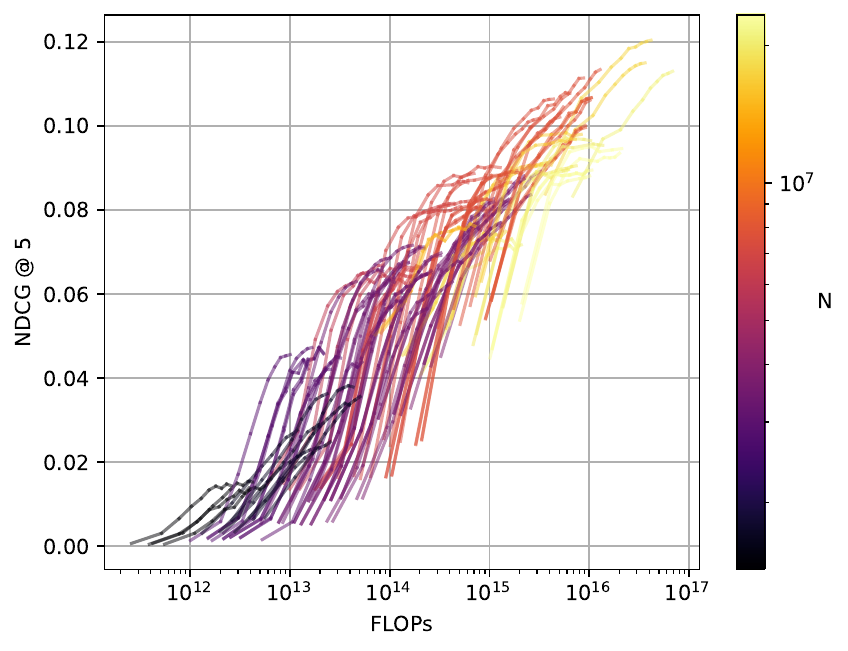}    
    \caption{NDCG@5 vs FLOPs for different runs with different training set sizes and model complexities. The colormap of each plot encodes the number of training interactions (left) and the number of non-embedding parameters (right). 
    }
    \label{fig:scaling_laws}
\end{figure*}

\begin{figure*}[h]
    \centering
    \includegraphics[width=0.45\linewidth]{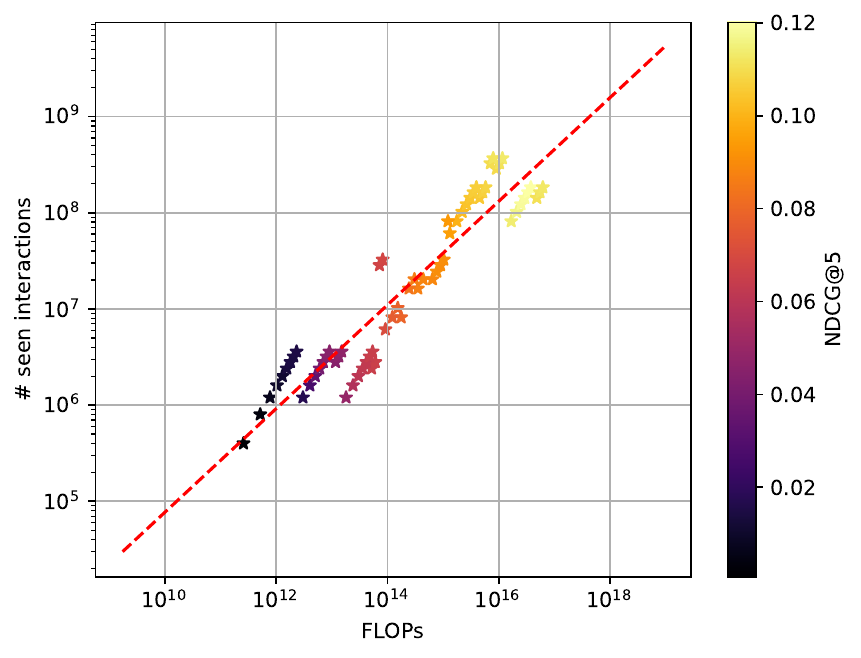}\hspace{15pt}
    \includegraphics[width=0.45\linewidth]{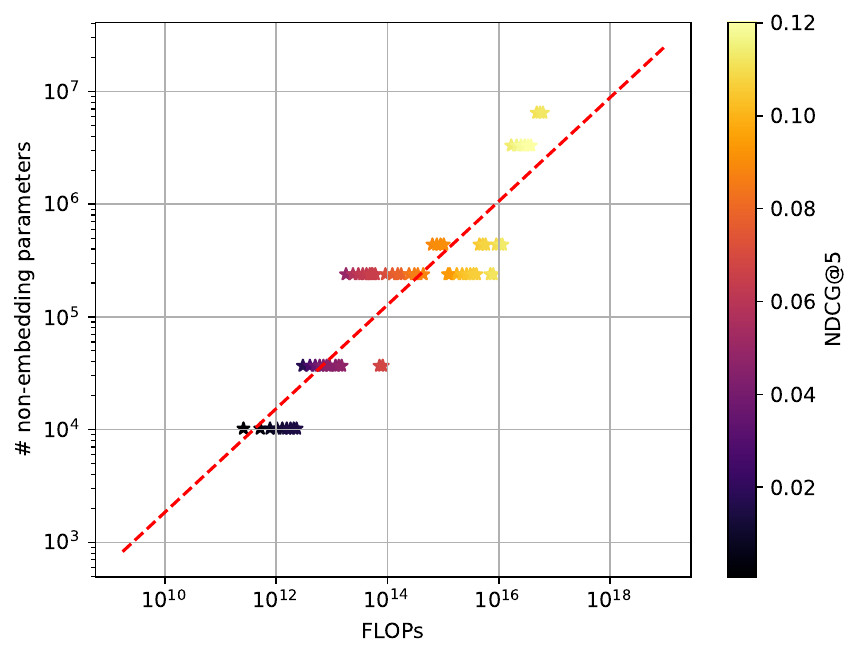}    
    \caption{Number of seen iterations (left) and number of non-embedding parameters as a function of the FLOP count for the points of maximal performance. Color encodes the NDCG@5 score of each configuration. The dotted red lines show linear fit curves of the corresponding point cloud in log-log space.}
    \label{fig:scaling_laws_envelope}
\end{figure*}

Figure~\ref{fig:scaling_laws} shows the target metric as a function of the number of FLOPs for different training runs obtained with different combinations in the number of model parameters and size of the training set. The left and right plots show the same runs but use a different color encoding to highlight different aspects of these runs. On the left, the colors encode the number of interactions processed in each training subset. This value ranges from $80K$ to $8.2M$. On the right, the colors encode the number of non-embedding parameters in each model. We decided to plot this number instead of the total parameter count for the following reasons: first, the number of token embeddings is constant across the different runs; second, non-embedding parameters (parameters of the transformer model) are responsible for capturing the sequential dependencies that are unique to our problem. The number of non-embedding parameters ranges between $10.2K$ and $9.6M$. The figures show an improvement in the target metric as larger models or bigger training sets are used. In smaller models, increasing the amount of training data reaches a point where performance saturates. Such data regimens are only useful if they come accompanied by an increase in the number of parameters in the model.
From these runs, we also extracted the envelope of maximal performance (\ie the points that observe the best NDCG score) among all the configurations that require the same amount of FLOPs. This envelope is highlighted in blue in the left panel of Figure~\ref{fig:scaling_laws}. From these points, we build scaling plots in Figure~\ref{fig:scaling_laws_envelope}. The plot on the left shows the number of seen interactions as a function of the number of FLOPs. The number of seen interactions is the number of interactions in the dataset used to train the model, multiplied by the number of epochs required to reach the point of maximal performance by the current configuration. We chose this quantity since we work on a multi-epoch setting where the optimal point for each FLOP count results from a model trained by a given number of epochs using a dataset with sequences of varying lengths. The panel on the right shows the number of non-embedding parameters as a function of the number of FLOPs for the points in the envelope. In both cases, color encodes the value of the NDCG@5 score. 

From these plots, we see a trend in that increasing the number of parameters in the model or the size of the dataset led to higher performance scores. Smaller models do not exploit the more significant variability observed in larger datasets. Reaching good performance by adding more data requires models with the flexibility to deal with this added complexity. In this case, however, it is not easy to disentangle the effect of increasing either of these factors. If we look at the figure on the right, we observe that a subset of models achieves different degrees of performance according to the resources devoted to training them (larger datasets or more training iterations). We believe, however, that identifying the scaling behavior brings valuable insights that allow us to extrapolate to novel data and model complexity regimes.

\subsection{Estimating Model Performance}
\label{sec:scaling_laws}

Based on the data obtained in our experiments, we present two formulations for estimating the expected performance in terms of the target metric for recommendation. These models aim at asking the following questions: \textit{a)} for a fixed FLOP budget, is it possible to get an estimate of the maximum achievable performance? and \textit{b)} for a given model and dataset size, is it possible to estimate the expected maximum NDCG for that configuration? In the first case, we assume there exists an "oracle" that selects the optimal model and dataset configuration.

\subsubsection{Estimating NDCG from a fixed FLOPs budget}

\begin{figure}[h]
    \centering
    \includegraphics[width=\linewidth]{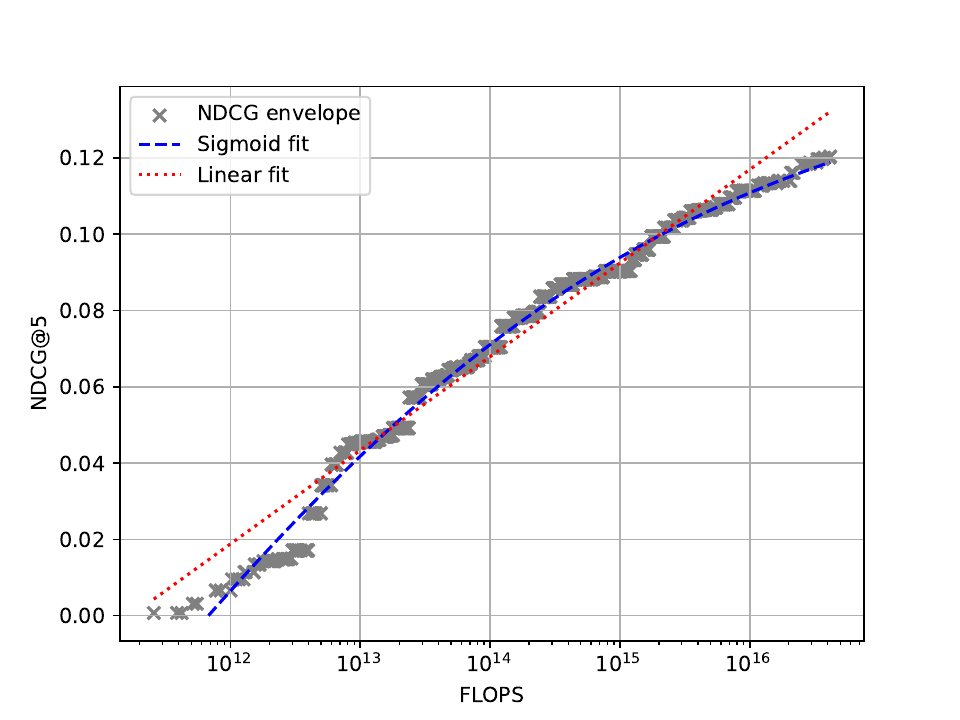}
    \caption{Relationship between the highest NDCG@5 values achieved at different logarithmically scaled FLOPs budgets. It includes two fitted parametric functions: a linear function and a sigmoidal function, showcasing the asymptotic trend of NDCG@5 as computational resources increase.}
    \label{fig:sigmoid}
\end{figure}
In the experiments, we recorded the maximum NDCG achieved for each FLOP budget. Figure~\ref{fig:sigmoid} shows such points together with linear and sigmoidal fits. The figure shows a more complex relationship between NDCG and FLOPs in log space than the linear scaling behavior observed in other studies. In our case, we observe the beginning of an asymptotic trend for the maximum achievable NDCG. This behavior could be due to many factors, including the saturation of the target metric due to challenges intrinsic to the particular recommendation problem (recommendation over broad item categories, representation ambiguity in the item embeddings, etc). In this case, a sigmoidal fit appears more appropriate, in which case it corresponds to:
\begin{equation}
   \text{NDCG}(FLOPs) \triangleq \frac{0.396}{1 + e^{-0.18(\log(FLOPs) - 24.44)}}-0.247.
\end{equation}
This function reveals that as the FLOPs budget increases, the NDCG approaches an upper limit estimated at 0.149 (0.396-0.247), highlighting the diminishing returns of increasing the computational budget. We can identify the point where this diminishing return starts at
$\log(\text{FLOPs})=30.7$, which corresponds to approximately \num{2.15e-13} FLOPs.
At this point, performance reaches a maximum estimated value of $0.0525$ $(0.155/2)$. Identifying the point of diminishing returns is essential in optimizing resource allocation in real-world scenarios.

\subsubsection{Estimating NDCG for a given model and dataset size}

\begin{figure}[h]
    \centering
    \includegraphics[width=\linewidth]{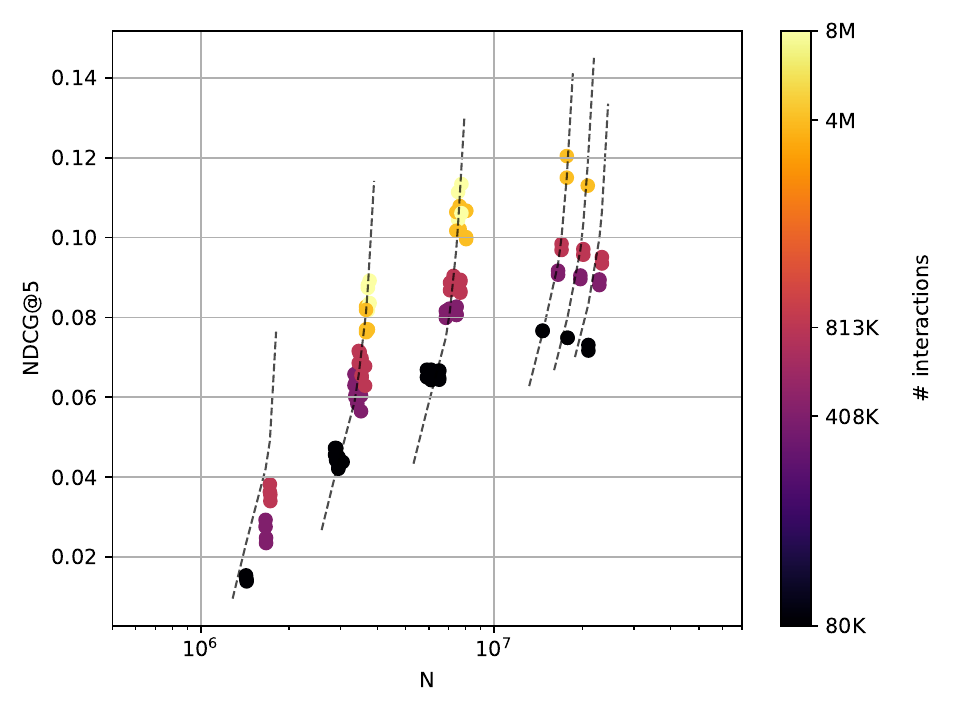}
    \caption{
    A view of the resulting parametric function from applying risk decomposition to NDCG values achieved at different model complexities (N) and size of the datasets as given by the number of user-item interactions.}
    \label{fig:ndcg_regimes}
\end{figure}

Here, we model the maximum achievable NDCG as a function of the total parameter count and the size of the dataset, as measured by the number of seen interactions. The goal is to find a parametric function that captures the underlying relationship between the model's complexity, data size, and final task performance. This involves identifying key parameters that influence the expected risk and then quantifying their impact on the model's effectiveness, as measured by the NDCG score. We follow a risk decomposition approach and propose the following functional form similar to \cite{hoffmann2022training}:
\begin{equation}
\label{eq:ndcg_n_t}
\text{NDCG}(N, T)\triangleq E - \frac{A}{N^\alpha} - \frac{B}{T^\beta}.
\end{equation}
Here, $N$ denotes the total parameter count and $T$ number of user-item interactions. We use a subtractive formulation to account for a target metric maximization law, instead of a loss minimization as in \cite{hoffmann2022training}.
The chosen parametric form allows us to outline how changes in the number of parameters and the size of the dataset systematically affect the model's ability to rank items accurately. To fit the model, we use a non-linear least squares approach and constrain the model coefficients to be non-negatives to avoid nonsensical solutions.
We obtain the following solution:
\begin{equation}
\label{eq:ndcg_n_t_sol}
\text{NDCG}(N, T)\triangleq 0.163 - \frac{18.56}{N^{0.376}} - \frac{2.9}{T^{0.364}} .
\end{equation}
From the above equation, we can interpret $E$ as the maximum expected value for the NDCG@5 score, in which case reaches a value of $0.163$. We also observe a similar value for the exponents for both $N$ and $T$, suggesting that both data and parameters behave similarly from a scaling perspective.
Figure~\ref{fig:ndcg_regimes} shows the predicted NDCG@5 score as a function of the number of model parameters, $N$, and number of seen interactions.

\subsection{Transferability}
\label{sec:transfer_learning}

In this section, we evaluate the transfer ability of some of our larger pre-trained models by fine-tuning them in the Amazon beauty and sports subsets. This is a widely used strategy in the literature but has seen lesser popularity in the context of sequential recommendation. This is because, unlike the language and vision domains, the data used to train such models are particular to each recommendation domain (\ie the nature of the catalog), and the type of events being recorded changes from case to case. Nevertheless, training more generic models at scale and fine-tuning them to different downstream tasks poses the same advantages observed in other domains, such as improvements in the final performance, shorter development cycles, improvements in the backbone model translate effortlessly to improvements in downstream performance, etc. We show that is it possible to pre-train and fine-tune recommendation models and that, by doing so, we obtain performance improvements that could not be achieved by training similar models from scratch.

Our adaptation strategy is as follows. Given a pre-trained model\footnote{To avoid data leakage, we removed the beauty and sports subsets from our pre-training datasets.}, we apply a progressive fine-tuning strategy that consisting of progressively unfreezing layer by layer, tuning them for 10 epochs using a learning rate of \num{1e-4} and a one-cycle cosine schedule. Once all transformer layers have been unfreeze, we unfreeze the token embedding layer and train for an additional 50 epochs. To avoid over-adaptation and catastrophic forgetting, we use the Elastic Weight Consolidation (EWC) formulation of \cite{kirkpatrick2017overcoming} which has been proven to be effective in ranking contexts \cite{lovon2021studying}. For EWC, we use $\lambda=100$.

Table~\ref{tab:fine_tuning} compares the results of the fine-tuned model, a similar model trained from scratch, and the following models from the literature: MINCE \cite{qiu2021memory}, ICLRec \cite{chen2022intent}, CoSeRec \cite{liu2021contrastive}, and 
CL4SRec \cite{xie2022contrastive}. Results are reported on the beauty and sports subsets of APD. We report NDCG@5 and HIT@5 scores. Table~\ref{tab:fine_tuning} shows that our fine-tuned variants outperform all alternatives by a margin. Interestingly, training from scratch on these datasets competes favorably with more alternatives from the literature, showing that despite its simplicity, our approach serves as a strong baseline for evaluations. If we look at the NDCG@5 score, we see an improvement of more than $12$ and $7\%$ for the Beauty and Sports subsets, respectively, for the fine-tuned variant \wrt to the models trained from scratch.
\begin{table}
    \centering    
    \caption{Performance comparison of our models with other methods from the literature.}
    \resizebox{\linewidth}{!}{    
    \begin{tabular}{lcccc}\toprule
    &\multicolumn{2}{c}{Beauty} &\multicolumn{2}{c}{Sports} \\
    \cmidrule(l{0pt}r{2pt}){2-3}\cmidrule(l{2pt}r{0pt}){4-5}
    &NDCG@5 &HIT@5 &NDCG@5 &HIT@5 \\\midrule
    MINCE \cite{qiu2021memory} &\underline{0.0378} &0.0523 &\underline{0.0196} &0.0274 \\
    ICLRec \cite{chen2022intent} &0.0324 &0.0493 &0.0182 &0.0283 \\
    CoSeRec \cite{liu2021contrastive} &0.0361 &0.0537 &\underline{0.0196} &0.0287 \\
    CL4SRec \cite{xie2022contrastive} &0.0208 &0.0396 &0.0116 &0.0219 \\\midrule
    \OURS-1K (from scratch) & 0.0360 & \underline{0.0611} & 0.0192 & \underline{0.0313} \\
    \OURS-1K (fine-tuned) &\bf 0.0405 &\bf 0.0645 &\bf 0.0206 &\bf 0.0344 \\
    \bottomrule
    \end{tabular}
    }
    \label{tab:fine_tuning}
\end{table}

\section{Related Work}
\label{sec:related_work}

Sequential recommendation is a branch of recommendation systems, an area that recognizes the importance of sequential behavior in learning and discovering user preferences \cite{wang2022sequential}. Initial models used the Markov Chain framework for anticipating user activities \cite{rendle2010factorizing,he2016fusing,he2017translation}. With advancements in deep learning, innovative approaches have emerged, such as employing Recurrent Neural Networks (RNN) \cite{hidasi2015session}, attention mechanisms \cite{sun2020go}, and Memory networks \cite{huang2018improving}. The disruption introduced by transformer architecture \cite{vaswani2017attention} led to significant progress, giving rise to well-known approaches like SASRec \cite{kang2018self} and BERT4Rec \cite{sun2019bert4rec}.
Despite their success, these methods face a substantial limitation in scaling. They rely on item IDs to represent the sequence of interactions, which presents several scalability issues \cite{fan2023recommender}. First, the pure ID indexing of users and items is inherently discrete and fails to impart adequate semantic information to new items. Second, adding new items requires modifications to the model's vocabulary and parameters, causing transformer-based methods to scale poorly with an increase in the item count, which is crucial for many real-world recommendation systems.

A viable solution to the constraints of ID-based recommender systems is to integrate textual information such as item titles, descriptions, and user reviews. The UniSRec model exemplifies this by deriving adaptable representations from item descriptions \cite{hou2022towards}. Text-based Collaborative Filtering (TCF) with Large Language Models like GPT-3 has demonstrated potential superiority over ID-based systems. Nevertheless, the overreliance on text prompted the development of VQ-Rec, which utilizes vector-quantized representations to temper the influence of text \cite{hou2023learning}. Additionally, approaches like ZSIR leverage Product Knowledge Graphs to augment item features without prior data \cite{fan2023zero}, and ShopperBERT models user behavior via purchase histories \cite{shin2021one4all}. IDA-SR advances this by using BERT to generate ID-agnostic representations from text \cite{mu2022id}. On the contrary, MoRec illustrates that systems that combine IDs and text can surpass those dependent solely on IDs \cite{yuan2023go}. However, these advancements complicate existing architectures by adding computational demand and complicating scalability.

To these intricate and parameter-intensive models, we must add the challenge that data in real-world applications is often noisy and sparse. Various methods have adopted contrastive learning \cite{oord2018representation} in new architectures, as seen with CoSeRec \cite{liu2021contrastive}, ContraRec \cite{wang2023sequential}, and S3-Rec \cite{zhou2020s3}. The success of these new contrastive learning-based methods motivates further investigation into the effectiveness of contrastive loss functions for item recommendation, particularly Sampled Softmax \cite{wu2022effectiveness}. Regrettably, these studies typically focus on fixed item spaces and overlook the scaling issues of the functions. Scaling problems have been tackled through other methods. LSAN suggests aggressively compressing the original embedding matrix \cite{li2021lightweight}, introducing the concept of compositional embeddings, where each item embedding is composed by combining a selection of base embedding vectors. Recently, the concept of infinite recommendation networks \cite{sachdeva2022infinite} introduced two complementary ideas: $\infty{-AE}$, an infinite-width autoencoder to model recommendation data, and DISTILL-CF, which creates high-fidelity data summaries of extensive datasets for subsequent model training.

Scaling issues are not unique to recommendation systems but are inherent in new transformer-based architectures. In the field of NLP, various studies have been carried out to discover scaling laws that predict the scaling of the model and inform decision-making \cite{hoffmann2022training}. To our knowledge, only two studies have attempted to find scaling laws in recommendation systems, yet none in SR. The first study \cite{ardalani2022understanding} aimed to explore the scaling properties of recommendation models, characterizing scaling efficiency across three different axes (data, compute, parameters) and four scaling schemes (embedding table scaling vertically and horizontally, MLP and top layer scaling) in the context of CTR problems. Similarly, the second study \cite{shin2023scaling} seeks to understand scaling laws in the pursuit of a general-purpose user representation that can be assessed across a variety of downstream tasks.

\section{Conclusions}
\label{sec:conclusions}

In this work, we studied the scaling behavior of the transformer architecture applied to real-world sequential recommendation problems. We introduced a simple and flexible architecture and learning formulation that allowed us to scale the recommendation problem and model complexity independently from each other. 

We showed there exist scaling laws similar to those observed in other sequential prediction domains, offering insights into the design of larger and more capable models. We also show that by pre-training larger recommendation transformers, we can fine-tune them for downstream tasks with significantly lesser data and obtain performance improvements compared to the same models trained from scratch. 

\bibliographystyle{ACM-Reference-Format}
\bibliography{sl.bib}
\end{document}